# HANDWRITTEN DIGIT RECOGNITION USING NEURAL NETWORK

by

ARKAPRABHA BASU

(Registration Number: 18370006)

**Project report Submitted in partial fulfilment of the requirements for the award of the degree of**

MASTER OF SCIENCE

in

COMPUTER SCIENCE

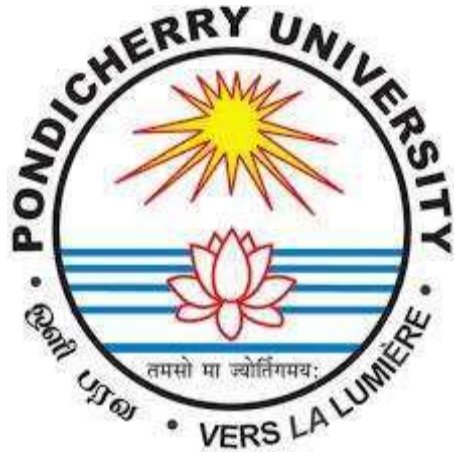

**DEPARTMENT OF COMPUTER SCIENCE**

**SCHOOL OF ENGINEERING AND TECHNOLOGY**

**PONDICHERRY UNIVERSITY**

**PUDUCHERRY-605014**

**MAY 2020**

# BONAFIDE CERTIFICATE

This is to certify that this project work entitled **"Hand-written Digit Recognition using Neural Network"** is the bonafide record of the work done by **Mr. Arkaprabha Basu (Registration No.– 18370006)** at the Pondicherry Central University in partial fulfillment for the degree of Master of Science in Computer Science, Department of Computer Science under the School of Engineering and Technology, Pondicherry University.

This work has not been submitted elsewhere for the award of any other degree to the best of our knowledge.

| INTERNAL GUIDE | HEAD OF THE DEPARTMENT |
|---|---|
| **Dr. M. Sathya** | **Dr. T. Chithralekha** |
| Assistant Professor | Professor and Head |
| Department of Computer Science | Department of Computer Science |
| School of Engineering and Technology | School of Engineering and Technology |
| Pondicherry University | Pondicherry University |
| Puducherry – 605014 | Puducherry - 605014 |

Submitted for the Viva-Voce Examination held on 07/07/2020

**INTERNAL EXAMINER**          **EXTERNAL EXAMINER**

(ii)

# ACKNOWLEDGEMENT

"No work can be completed successfully without the proper guidance & help
Of the trainer and other people."

This project report is also a manifestation of the invaluable guidance, suggestions, support and encouragement extended by various people.

My sincere thanks to my project guide Dr. M.Sathya for guiding me. Our special thanks to our university department head Dr. Chithralekha for providing us with this wonderful opportunity and for their valuable guidance and moral support.

I am very thankful to our project guide Dr. M. Sathya for helping me to dive into the project every time. I would also like to thank Dr. T. Vengattaraman who helped me in understanding and visualizing the project in a better way. Their regular guidance gave us the direction to work during the project.

I would also like to thank my parents for all the support and strength. I am very grateful for our friends for helping us to get into this topic as my project.

I am very thankful to all the other staff members, librarian & other co-workers for their help in the completion of this project.

# SYNOPSIS

The project work is a practical experience of the knowledge one has.The documentation leads a way to the concept to present the thinking and the upgradation of various techniques into the project .This project entitled "**HANDWRITTEN DIGIT RECOGNITION**" is a practical project based on some trends of computer science.Every day the world is searching new techniques in the field of computer science to upgrade the human limitations into machines to get more and more accurate and meaningful data.The way of machine learning and artificial intelligence has no negative slop it has only the slop having positive direction.This project is a very basic idea of those concepts .This project deals with the very popular learning process called Neural Network. There are various ways by which one can achieve the goal to a desired output,but in machine learning Neural network gives a way that machine learns the way to reach the output.

This project has come through the concepts of statistical modeling,the computer vision and machine learning libraries which includes a lot of study about these concepts.I tried to lead these project to the end of some updated techniques,upgradation and application of some new algorithms . This project has a good explanation and this project can be enhanced further into some complex applications of machine learning.

# TABLE OF CONTENT

# LIST OF FIGURES

# CHAPTER 1

# INTRODUCTION

## 1.1 ABOUT THE ORGANIZATION

### PONDICHERRY CENTRAL UNIVERSITY

Pondicherry University is a Central certified University in Pondicherry the place of Sri Aurobindo and sea, Kalapet, R V NAGAR. Just coming from the Pondicherry town one will find it on the left side after the stoppage of Pondicherry Engineering College.This University is stated as a rank of excellence and government funded institute which plays a key role to teach and lead the students up to excellence.This university is situated with 53 departments including science,arts and commerce.The foreign language in various departments has some training and paid courses also.This university has some advanced lab systems and facilities in various science departments where PhD scholars,teachers and students are always welcome to practice for the excellence of their various practical results.

The respective Head of the department along with the other respective faculties provide the Department of Computer Science, which is under the School of Engineering and Technology a center of excellence for the students.The laboratory contains newly upgraded machines with i5 processor and good internet facilities and server system. During the course there were opportunities to know so many new things and upgraded techniques and along with the project and lab work, there was provision to chance to practice one's excellence. The faculty advisor of the department at first described the course structure to the students as some of the students didn't know the CBCS (Choice Based Credit System).

This institute is strong and has a very pleasant hostel life, a quite advanced library and reading room within .The students always got the help from the upper admin side which makes the life easier and for which they can concentrate in their study and research.

**1.2 ABOUT THE PROJECT**

The project comes with the technique of OCR (Optical Character Recognition) which includes various research sides of computer science .The project is to take a picture of a character and process it up to recognize the image of that character like a human brain recognize the various digits.The project contains the deep idea of the Image Processing techniques and the big research area of machine learning and the building block of the machine learning called Neural Network.There are two different parts of the project

1) Training part

2) Testing part.

Training part comes with the idea of to train a child by giving various sets of similar characters but not the totally same and to say them the output of this is "this". Like this idea one has to train the newly built neural network with so many characters.This part contains some new algorithm which is self-created and upgraded as the project need.

The testing part contains the testing of a new dataset .This part always comes after the part of the training .At first one has to teach the child how to recognize the character .Then one has to take the test whether he has given right answer or not. If not, one has to train him harder by giving new dataset and new entries. Just like that one has to test the algorithm also.

There are many parts of statistical modeling and optimization techniques which come into the project requiring a lot of modeling concept of statistics like optimizer technique and filtering process, that how the mathematics (*How to implement a neural network intermezzo 2 Peter Roelants (2016)) and prediction (Kaiming He et al))* behind that filtering or the algorithms comes after or which result one actually needs to and ultimately for the prediction of a predictive model creation.Machine learning algorithm is built by concepts of prediction and programming.

# CHAPTER 2

# PROBLEM DEFINITION AND FEASIBILTY ANALYSIS

## 2.1 PROBLEM DEFINITION

The total world is working with the various problems of the machine learning.The goal of the machine learning is to factorize and to manipulate the real life data and the real life part of the human interaction or complex ideas or the problems in the real life.The most curious of those is Handwritten Character Recognition because it is the building block of the human certified and the classification interaction between other humans.

So, the goal was to create an appropriate algorithm that can give the output of the handwritten character by taking just a picture of that character. If one asks about Image processing then this problem can't be solved because there can be a lot of noises in that taken image which can't be controlled by human.The main thing is when human write a handwritten character or for our case digit he has no single idea whether he has to draw it in the circulated pixels or just same as a standard image given .A machine can do that but not the human.So by matching only the pixels one can't recognize that.

The idea of machine learning lies on supervised data.Machine learning algorithm fully dependent on modeled data .If someone models the Image directly, the model will get a lot of flatten values because that picture can be drawn with various RGB format or with various pixels which can't be modeled accurately due to noise.

So, for this project one has to create a model by image processing and the machine learning. Both the techniques will be needed because these two techniques will enhance the technique of the machine learning and that can shape this project.

## 2.2 NEED OF THE SOFTWARE

The total project lies with a great computation speed and by a online server where run and compilation done quickly. All the packages were imported that were needed for the software online. We need the tools to be imported also.

This project at first is in need of the software of python.The total code is written in python so it needs Python3. Python2 was not chosen because python3 has some additional upgrade over python2. The packages have been imported and the algorithm created which is done by installing the new packages from online in python3.

Apart from that the total project is online compiled or ran and done by the software provided by the Google Colab free version.Apart from choosing Anaconda Navigator, Spyder or Jupyter Notebook, Google Colab or Colabotory have been chosen because this provide more speed and accurate compilation as is known. Creation of the machine learning algorithms deals with data and bigger size programs.

## 2.3 BETTER FUNCTIONALITIES

This project deals with the End users with some functionalities.The major functionality can be the Recognition of digit.User can write a handwritten digit and this project will recognise it accurately. Edge detection can be set in the process of image processing.ML algorithm can differentiate the various digits from another by recognising it.

The better functionality where the building block can be this project is Mathematical Model solver. One can take any picture of a mathematical problem and by this project one can recognise the digit inside it and then computer can compute that problem on its own.If a wrong answer comes, it can be checked through a step by step process by the computer and if it recognized the answer wrongly, it must be trained again. One has to train the model in various extents to recognise the various digits not only 0 to 9 but also more and more figure, like derivative integration and others.

The better functionality of this project can be license plate verification.Car license plate can be checked and one can set the record rightfully that which car is passing the gate and when by the recognition of characters.

## 2.4 FEASIBILITY ANALYSIS

There are a lot of sides of feasibility. Let it be discussed one by one.

### 2.4.1 Technical Feasibility

The software which has been used in this project is fully open source and one can connect to it whenever he or she wish. The concept of python and open CV another side the research concept of image processing and machine learning is a very trending topic nowadays .Apart from that all the software running environment Google Colab is fully open source and easily can be accessed in the presence of internet.The user can also be a non-programmer and by clicking the run button he or she can set the digit in the webcam screen and can see the output.

### 2.4.2 Seasonal Feasibility

This project is feasible in time that means this project has been started in a particular date and completed in the available time fully. It was an efficient effort which resulted in completion of the project in time.

### 2.4.3 Economic Feasibility

This project is economically free because all the open source softwares have been used that is why no money was charged or given.Only the study materials in the devoloper side or the designer are not available in free of cost.The software or the program is fully free of cost.

### 2.4.4 Profitability

This project deals with two trending topics.Image Processing and Machine learning.The total machine learning concept has not used here but we have dealt with the building block of the machine learning called Neural Network. These two topics are an appropriate research topic and many scholars and teachers also are working with it every day to upgrade the techniques or the algorithms or to create some new algorithms. The extension of the project can be used in large scale to detect the written character from images and to extract them in no time.Or in banking signature recognition or in license plate verification, Real time image chaining or filtering or object detection etc.This project is fully profitable because many sides of this project is in research nowadays and government is funding those researches.

# CHAPTER 3
# SOFTWARE REQUIREMENT SPECIFICATION (SRS)

## 3.1. SYSTEM REQUIREMENT

This project needs the help of hardware and software requirements to be fit in the computer or the laptop Pc.The user and the toolkits and hardware and software requirements are required also.

### 3.1.1 Hardware Requirements

RAM: At least 4 GB.

Processor: Intel(R) core (TM) i3 or more. 2.00 Ghz.

Internet connectivity: Yes.(Broadband or wi fi) Webcam

connectivity: Yes

### 3.1.2 Software Requirements

| SOFTWARE | TYPE/PLATFORM | VERSION |
|---|---|---|
| Operating System | Windows, Linux ubuntu,fedora or similar | Windows 7 or more, ubuntu 16 or above, fedora 20 or above |
| Python | -- | Python 2.7 or above |
| Opencv | -- | Opencv 3.2.0 or above |
| Numpy | -- | With Python 2.7 3.5 |
| Tensorflow | -- | Tensorflow 2.1.6 or above |
| Ipython | -- | Ipython 7.10 or above |
| Pil | -- | Pil 1.1.5 or above |

## 3.2 TOOLS AND TECHNOLOGIES

### 3.2.1 OpenCV

The Opencv(Open Source Computer Vision Library) is a python DIL library which is very updated in the work of Image processing .One image file and pixel values can easily come into surface by this library.This library provides a common infrastructure and module related to computer vision technologies.The most important thing about this tool is it is totally free and can be easily modified and changed respective to input by the programmer.

### 3.2.2 Numpy

Numpy is a library for the python programming language, adding support for large multidimensional array and metrices.This package contains a large collection of high level mathematical functions to operate on those arrays. Numpy is open source and has many contributors.

### 3.2.3 Tensorflow

TensorFlow is an end-to-end open source package in python for machine learning. It has a comprehensive, flexible ecosystem of tools, libraries and community resources that lets researchers push the state-of-the-art in ML and developers easily build and deploy ML powered applications.

### 3.2.4 IPython

Ipython is something different .It provides a rich architecture for interacting computing with:

    A powerful shell

    A kernel for jupyter

    Support for data visualization like matrix values flexible  interpreters for easy use.

### 3.2.5 PIL

PIL is a library of advanced image tools having full name Pillow.It is used for noise cancellation and to draw modified pixels by noise.It is useful to Image crop and various subjectial techniques.

# CHAPTER 4
# MODEL DETAILS

**4.1 IMAGE PROCESSING**

Image processing technique has been implemented extensively at the very first part of the project.

So, what is Image Processing? Image processing is a method to perform some operations on an image, in order to get an enhanced image or to extract some useful information from it. It is a type of signal processing in which input is an image and output may be image or features or characteristics associated with that image. Nowadays, image processing is among rapidly growing technologies. It forms core research area within engineering and computer science disciplines too.

Image processing basically includes the following three steps:

1) Importing the image via image acquisition tools
2) Analysing and manipulating the image
3) Output in which result can be altered image or report that is based on image analysis.

In this project at the last part of detection to take the input directly from the webcam, to reshape that image and in the very first part of training MNIST dataset image reshaping and real life dataset written reshaping, cutting, filtering all requires the idea of the image processing. One by one the concepts of image processing in this project will be covered. As this project remains in two parts the impact of Image processing will be discussed in two parts

1) Image processing in training data.
2) Image processing in testing data.

**4.1.1 Image processing in training data**

In the training dataset the neural network model has been trained with two different dataset.
i) MNIST (Modified National Institute of Standards and Technology database)dataset
ii) Dataset self created.

This two dataset almost set 60500 entries in the training model and trains the model with excellence almost 5 times for iteration.

i) The MNIST database (Modified National Institute of Standards and Technology database) is a large database of handwritten digits. The database is also widely used for training and testing in the field of machine learning. It was created by "re-mixing" the samples from NIST's original datasets. The creators felt that since NIST's training dataset was taken from American Census Bureau employees, while the testing dataset was taken from American high school students, it was not well-suited for machine learning experiments. Furthermore, the black and white images from NIST were normalized to fit into a 28x28 pixel bounding box and anti-aliased, which introduced grayscale levels.

ii) Handwritten characters self- created this dataset almost adds 500 entries in the training of the dataset. This is more efficient than MNIST dataset for this project and serves at the last part of the training model as it can face more and more problem like this after for testing. This dataset has its own characteristics

1. Has no fixed value of the image.
2. Mobile picture taken from a paper
3. Multiple characters are given and then cut by the algorithm
4. After cutting and resizing process saving it to the particular directory.

This dataset require a lot of Image processing works before sending into the neural model. There are various unlabelled factors in this image or dataset which needs to be controlled.

1. Multiple characters are given at random, need to cut them.
2. The real time mobile image can be of different shades.
3. Noisy image will come up; we need to make it noiseless.
4. One needs to resize the image depending on the quality loss.

Steps included into the problem in Table 4.1 as mentioned below as the various parts of Image Processing parts on the training data taken by self created dataset.

Table 4.1 Image Processing parts

| MOBILE PICTURES OF CHARACTER WRITTEN ON A PAPER |
|---|
| VARIOUS ALGORITHMS APPLIED FOR NOISE FILTERING |
| CONTOUR DETECTION |

**4.1.1.1. Mobile picture of characters written on a paper**

This step includes some writing of characters into a black and white paper because that will be more efficient to detect, unless one can write in margin based page also.Writing with sketch pen or marker or gel pen will be very nice because those create continuous characters.As ball pen or pin point pencils can create disconnected components within a character those can't create good character and good character writing are necessary while it is needed to scan it and want to recognise it by machines for OCR.

After writing in paper go through the process of taking the picture.This process is simple, no scanning is needed ,only take a good picture of the characters.

**2. Various algorithms applied for noise,filtering**

This step is fully localised on cleaning the data,removing the noises and filter the data to get the actual subjectial data from the image.

**2.1. Normalisation**

Real world data stays in various forms and if a scatterplot of the black white variance is created it will give median filled results also.That is why normalisation is necessary to get the accurate cleaning process.

Here the MINMAX Normalisation(eq 4.1) technique has been used to standardise the image. Here the picture range of colors has been removed and a range was given for each and every pixel like 0 and 1 (*Peter Roelants – 2016*).

$$V_i{}^` = (v_i - minA)*(newmaxA - newminA)/(maxA - minA)$$

Equation 4.1 Minmax normalization

where $v_i'$ is new normalised pixel number, $v_i$ is old pixel number(range minA to maxA) and the newminA to newmaxA is the range of the new pixel.

There are many types of normalisation techniques but minmax was chosen because it is tested to be best for the handwritten character cleaning. It glows the black pixel to black and the noisy pixels (gray) to white more efficiently.

**2.2. Filtering**

Filtering is the way to smoothen or sharpen an image.This process is done by removing very high or very low frequency cells.Low pass filters remove the higher frequencies by keeping only the lower frequencies.The work done by the low pass filter is to smoothen the picture and remove the noises.High pass filters remove the lower frequencies by keeping only the higher frequencies.The work done by high pass filter is to sharpen an image with no background(with black background).As the low filter used in noise removal, we have implemented it first.

1. To remove more noises on the first phase, high pass filter and applied pyrmeanshiftfiltering() are suggested. In this, the initial step of mean shift segmentation of an image is carried out.Pyrmeanshiftfiltering is Pyramid Mean Shift Filtering on an image using Pyrdown() method of the imgproc class.

$$pyrMeanShiftFiltering(src,dst,sp,sr)$$

src- Source Image
dst- Destination Image
sp- Type double representing spatial window radius
sr- Type double representing color window radius

$$filteredimage=cv2.pyrMeanShiftFiltering(img,21,111)$$

This filter removes all the boundaries like the folded side margin of notebook and other dependencies as this is segmented so it glows the black part of that image also.

2. To remove more of the noises on the second phase low pass,the medianblur filter are applied. This part the following Table 4.2 removes the paper noise from the image caused by the paper only.This type of noise is called Salt and Pepper Noise.

Table 4.2 Type of Noises

Pepper Noise

| 0  | 30  | 45 |
|----|-----|----|
| 50 | 100 | 55 |
| 60 | 10  | 90 |

Salt noise

To remove the Salt(High frequency) and Pepper(Low frequency) noises use the median filter.So,the intensities in the example image converted into median f(x.y).

$$f(x,y)=\{0,30,45,50,100,55,60,10,90\}$$

increasing order $f(x,y)=\{0,10,30,45,\mathbf{50},55,60,90,100\}$
_________Median

$$median\ f(x,y)=\{50,50,50,50,50,50,50,50,50\}$$

This process continues through neighbourhood processing for the whole image and what size is given, the matrix will be taken in that way only.This medianblur process kept on the second phase because if this process applies on the first as this is a low pass filter it can remove the main subject also.So,the background line is removed first and then applies it to keep the main subject and only to remove salt and pepper noise.

3.The Thresholding is done on the third step due to some new advancements of noise cancellation and dot taking technique.Salt-Pepper noises and boundary margin noises can't give the solution of dot noises.Those noises are taken by the technique of the bounding box category by taking the pixel for each and every character.

$$matrix(x,y,width,height)$$

where, x,y is the point in the 2D plane and width and height is the dimension of the bounding box.

4. The effort is to make the total image noiseless from the very first and now that is very close to the aim. Here comes the time to see how much the effort have achieved the goal by applying the high pass binarization canny filter on the image. The canny filter is one type of filter that determines the edges of the variance of a grayscale image and considers it to be a border. This filter is applied in a modified way that it can glow the border in 0 like white and 1 like the background. Maxval,minval is the *range* of the image. Third argument can be aperture_size.

$$cannyimg=canny(img,minval,maxval)$$

**3. Contour Detection**

The contour detection technique is fully applied after the noise cancellation of the image, after normalisation and the filtering process. This process comes with some real issues which can happen only for handwritten characters. Thereare various parts of contour detection in this paper. One shall be dealing with various parts one by one and solve the real issues that are faced in case of contour detection. Contours are some curves or lines joining the same color or intensity points along the boundary. This is a very much useful tool for object detection and recognition which takes the subjected shapes from an image. For better accuracy in contours, binary images or canny image must be taken where subjected shapes and background are widely different intense.

Mentioned Table 4.3 depicts some steps on the contour upgradation algorithm.

Table 4.3 Contour detection upgradation

| Hierarchy |
|---|
| Bounding Box |
| Putting Contours |

### 3.1. Hierarchy

At first,to know the contour hierarchy,how to find contours of a binary image?

region,hi=cv2.findContours(edged.copy(),cv2.RETR_TREE,cv2.CHAIN_APPROX_SIMPLE)

There are three arguments in cv2.findContours(); first one is the source image; second is contour retrieval mode; third is the contour approximation method.

Contour hierarchy deals with contour retrieval mode and contour approximation method. There are various types of retrieval and approximation method.

findcontours() retreives all the contours in an image,but sometimes or most of the times for handwritten character set some contours stays inside one contour or connected with another contour.The outer contour called Parent,inner connected contour called Child and the connected contours called Sibling.This type of relationship is Hierarchy of contours.

Fig 13 and Fig 14 depicts the hierarchy between contours. Digits 6 and 5 is determined by two contours one parent (bigger one), one Child (smaller one). Digit 8 is determined by one parent and two child contours,the relation between children can be called as sibling.

There are many methods or functions in opencv which can determine contour hierarchy so clearly.

    cv2.RETR_LIST(): Parent and kids belong to the same hierarchy level.
    cv2.RETR_EXTERNAL(): Only parent is taken care of no other members.
    cv2.RETR_CCOMP(): Only takes hierarchy 2,members,not parent.
    cv2.RETR_TREE(): Take out the total family,grandpa,son,grandchild all.

*Problem of Contour Hierarchy*

Now the main theory about contour is the contour is totally dependent on variance on the pixel,like whenever it gets a variance it applies the contour on it. So, in case of handwritten digit extraction it's not efficient because inside an image it will consider more and more sibling contours and then total family while only the supreme of that family is needed.This thing totally can't be considered because inside one contour one child is there but comparing to another it can be parent also.Inside one contour there comes another contour also that will extract more than one contour from a character.in fig 6,for "Q" it will cut 4 contour 1 outside and three inside contour.For "8" it will cut the same also.This is problematic.

**3.2. Upgradation in Bounding Box**

In this paper, the problem in the concept of bounding box will be upgraded.Mainly, Bounding box is one type of box which can be rectangle or oval shaped or circle by which an object in a particular image or video can be recognised.Mainly, for handwritten character recognition rectangle boxes can be used to describe the target location box.Bounding box gives the output in a 4 valued list. See fig 5 for more details.

$$x,y,width,height=[162, 191, 74, 74]$$
$$A(x,y) \text{ upper left point of the contour.}$$
$$B(x,y-height) \text{ lower left point.}$$
$$C(x+width,y-height) \text{ lower right point.}$$
$$D(x+width,y) \text{ upper right point.}$$

**3.2.1 Matrix insertion**

So what has to be done is to delete those bounding boxes which are inside and to keep only those bounding boxes which are outside and giving the boundary to the characters. For Fig 12, 13 only the outside box has to be kept and not the inside. All the bounding boxes values can be obtained and put into one matrix or 2-D list.

**3.2.2 Duplicate or Bounded bounding box deletion**

There can be many bounding boxes which are fully bounded by another bigger bounding box that need to be deleted because there can be pixel variation inside a character but that will not be considered.Only outside bounding box will be considered.

The picture has the upper left point(x,y),width and height of every bounding box has been stored in our matrix.Take every two lists from the matrix and check if those bounding box has any bounded bounding box or not.

Check if the one bonding box (x,y) point inside another .If inside then check the width+x and height+y if inside bounding box is fully inside.If yes then delete it or add another one into mainmatrix.After all take only distinct values. Then sort the matrix as those contours need to be obtained and to fit it into the distinct positions.Like the contour which has been created for 8(fig 12) is residing in a matrix so we have to fit that contour onto 8 only and same for another characters also.If one bounding box is started inside one but its height or width goes beyond the bigger one which is mostly outside then inside contour will be there.Only fully bounded inside contour will be deleted or not been considered.

**3.3 Putting contours**

After detection of the right contours for the image, in the ideal contour bounding box recognition the number of bounding box found will be equals to the number of characters in the image. After getting the bounding boxes it is needed to be sorted with removing the duplicates and then those bounding boxes on the image setting contours have to be put.

$$x,y,w,h =matrix[i]$$
$$imag = cv2.rectangle(image,(x,y),(x+w,y+h),(0,0,255),2)$$

The very first attribute on the rectangle function is the image on which one will draw the boxes. Second attribute (x,y) is the upper left corners of the contour in that 2-D image.Third attribute will denote the right side bound and the lower bound of the image.Fourth attribute will be the color of the contour box (r,g,b) and the fifth one will be the thickness of the bounding box.

## 2. Results

The following Table 4.4 depicts the result of the contour removal normal algorithm and updated contour removal bounding box based algorithm. The accuracy on the new algorithm has been calculated and also given there.

Table 4.4 Total upgradation in results after applying bounding box removal

| Total Tested Samples | Contour Removal Algorithm | Updated Contour Removal Bounding Box | Accuracy |
|---|---|---|---|
| 134 | 56 | 119 | 88.8 |
| Noise Cacellation | | | |
| 120 | 110 | 116 | 96.6 |

Now after getting all of these results the training data gets trained with the total MNIST dataset and the real dataset into the neural network model.

### 4.1.2. Image processing in Testing Data

In the case of testing data at first I have thought it will be a picture which will be already in the database written by a human.To make the testing part more interesting, real time photo is captured by the webcam of the computer.This part has increased the complexity of the program and also of the project but made it more interesting.And then the three parts like the training dataset handwritten digit removal has been applied on those.In this process there are two major parts .

1) Accessing the webcam and capture the picture

2) Noise cancellation and resizing and scaling.

*1) Accessing the webcam and capturing the image*

In this part google colab is taken as interactor software with webcam .Used the IPYTHON and javascript in colab to set the fitted image into the webcam.Javascript will start the video it will show whatever comes into the mirror camera.The stream will be constant until the capture button comes 'True'. If capture comes true the designated pixel will be counted and will be stored as photo.jpg in the internal storage of the google colab.In IPYTHON display a 2D array is manually set to get the image pixelated input into the image.The capture is set as a promise function.

2) *Noise cancellation and resizing and scaling*

From the Ipython display the value of the pixels has been copied to a numpy array and then using a 20,20 kernel the image has been designed. The normalisation of that image is done as previous training dataset entry and scaling and resizing done as the same.Pyrmeans filtering have been removed because the picture which has been taken in the webcam has been transmitted through the numpy kernel at first as the picture is of the single character only.

The Image processing part is very important for this project because the total preprocessing of a machine learning algorithm lies through the part of the Image processing.The total unsupervised to supervised transformation can be done with the help of image processing. In the project, the preprocessing work concludes webcam access, inputs of the pictures, scaling, resizing, filtering, noise cancellation and at last sending to the machine learning algorithm as a kernel for testing.

## 4.2. THE NEURAL NETWORK MODEL

The neural network lies with the concept of machine learning.At first it creates a model like brain unit and then like child it trains that model like brain with many and many datasets,for my project it is the digits.There are two parts of a neural network model.

1) Training Part
2) Testing Part

Before going into the deep concept of these two parts in the project some ideas on the building blocks and structural parts of neural network (*C.C.Jay.Kuo – 2016, Adit Deshpande – 2016*) have been shared. Each and every neural network can be structured in three different parts-

1) Input layer
2) Hidden layer
3) Output layer

A neural network is put together by hooking together many of our simple "neurons," so that the output of a neuron can be the input of another. For example, here is a small neural network:

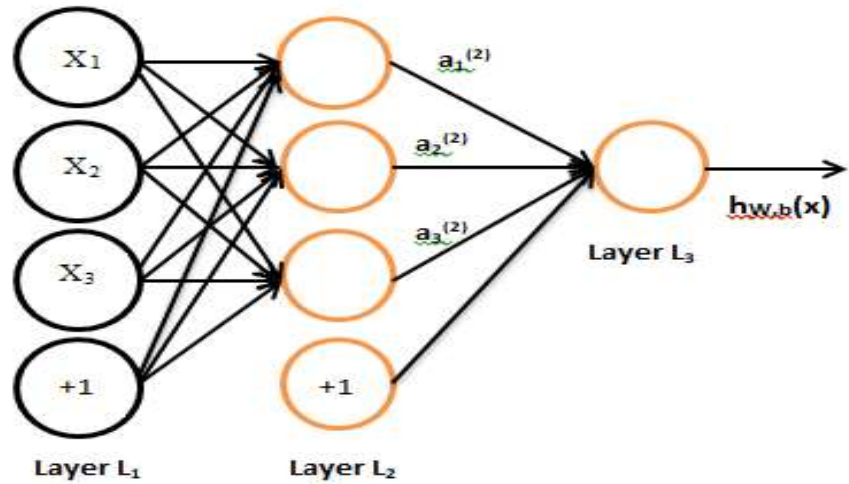

Fig-4.1 Neural Network Model

In this figure 4.1, circles are used to denote the inputs to the network. The circles labeled "+1" are called bias units, and correspond to the intercept term. The leftmost layer of the network is called the input layer, and the rightmost layer the output layer (which, in this example, has only one node). The middle layer of nodes is called the hidden layer, because its values are not observed in the training set. Say, that our example neural network has 3 input units (not counting the bias unit),3 hidden units, and 1 output unit.

The types of layers in a neural network can be summarized as follows:

*1) Input Layer*

Input variables, sometimes called the visible layer.This layer can be of features or the direct input from the dataset.

*2) Hidden Layer*

Layers of nodes between input and output layers . There may be one or more of these layers.More one create the hidden layer more operation will be there,more weights will be there and there will be less change due to the value of difference in this layer.This layer is the most important part of the neural network.Weights change in these layers varies in the returning input.

*3) Output Layer*

A layer of nodes that produce the output variables and then based on the targets it can perform iteration.Finally, there are terms used to describe the shape and capability of a neural network; for example:

Size: The number of nodes in the model.

Width: The number of nodes in a specific layer.

Depth: The number of layers in a neural network.

Capacity: The type or structure of functions that can be learned by a network configuration. Sometimes called "representational capacity". Architecture: The specific arrangement of the layers and nodes in the network.

## 4.2.1 TRAINING PART

In the training part of the neural network a neural network model designed with various factors and determining other statistical learning model that can differ based on the dataset given has been created.This neural network then can be trained with a lot of datasets (for our case 60500 datasets) for better results and to upgrade the weights into the neural network.

## 4.2.1.1 CREATING THE NEURAL NETWORK MODEL

Now comes the main part of the project ,so how to create a neural network model that can take an image and run some probabilistic calculation on it which can implement the model with better accuracy.

*1) Input Layer*

The total model trains on the dataset 28*28 pixels .The neural network model depends upon the number of inputs and the intensity in input layer and the filtered target output calculation in output layer. One dimensional array of the intense dataset is needed into this model and then sequentially give the total MNIST dataset and out retrieved dataset into the neural. One needs to flatten and sequential the total input into the neural.

*2) Hidden Layer*

The hidden layer part is the most efficient part in the neural network.It is the part where the total calculation and the backpropagation like learning algorithm takes place.Inthis project 128 hidden layers have been applied by taking the data from the input layer. In inner input layer and dense hidden layer MAXRELU function has been applied as an activation function and in the output layer the softmax activation function has been used (*Softmax output function, Jefferey Hinton*).

*What Activation Function actually does?*

Activation function controls the boundary of the data like normalisation .The normalisation function are rejected in the hidden layer because in the hidden layer the data or the values gets dynamically changed.There are various types of Activation functions.

a) Sigmoid- $Y=1/(1+\exp(-x))$

b) Tangent- $y=(\exp(x)-\exp(-x))/(\exp(x)+\exp(-x))$

c) MaxReLu- $y=\max(0,x)$

d) Softmax- $y=\exp(x)/(\text{summation}(\exp(x)))$ etc.

In this project the max relu and softmax because this gives really good result tested and better than others like fig 4.2.Total experiment is based on result no extra theory applied but the convergence of the sigmoid and the tangential curve to the verified output scenario.

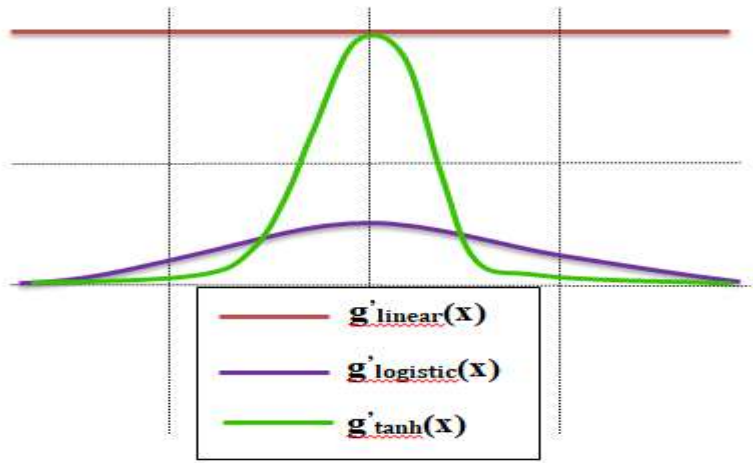

Fig – 4.2 Types of activation functions over graph

As the linear,logistic(sigmoid) and tangential is too much deflected for stagnant inputs and touches the curve in 1 and 0 immediately and never comes back practically ,these 3 are rejected.Sigmoid gives a very good result but practically max relu and softmax is better for Handwritten digit recognition.

Bias input are rejected to contain the actual shaped data from the main flatten and sequential data. Learning rate has been given very low 0.002 to make the model more effective.

*A) OPTIMIZER*

Optimizers are algorithms or methods used to change the attributes of the neural network such as weights and learning rate in order to reduce the losses.

There are many types of optimizers.Each and every optimiser has its own method to deal with the weights and bias inputs.In this project the backpropagation algorithm is modified with the adam optimizer instead of the gradient descent optimiser. At first gradient descent was set to be final but then the adam or stochastic gradient descent optimiser is better than anyone as the inputs from the various pixels are varying in points.

**4.2.2. TESTING PART**

The testing part comes with some latest real time update in the machine learning part .The image processing work like webcam access and the other techniques has been discussed before that is why it has not been discussed here. So this part comes with some ransom upgrade of the technique which is very small to say but not that much easy to design.

Let's dive into the model after the total training and taking the figure from the webcam into that predictive modeled neural network. Set the input pixel values into the numpy array value where the argmax function will check the range and give the output figure .The range will be done by the feed forward step of the neural network.No other backpropagation algorithm will be applied because this is the test to check have the weights updated them to detect that input figure correctly? NO neural network can come with 100 % accuracy but if it can't detect most of the test figures then it must need more training. Result of the model shall be discussed in no time because the model has been trained more than 23 times to make it better and of course not with the same dataset every time. Datasets include ones from web and pictures and also handwritten and trained the neural model to give the correct result in the testing part.

To make things shorter what has been done in this model-

1) Loaded self created data and MNIST dataset into the model
2) Set the numpy array system to take input the kernel along with the data.
3) Create the neural network model setting the input layer and the number of hidden layers and the output layers along with the activation functions used in different layers.
4) Set the probabilistic statistical value into the biased dataset into unbiased.

5) Check the target and the output predicted value every time while training the dataset and set the number of epochs corresponding to the error.
6) Model check with checksum value removal and biased value removal after setting the weight value by the neural network on its set and set the dataset into the unbiased.
7) Access the webcam and capture the image using python javascript and google colab display.
8) Check the parameter of probability of the figure after normalisation and noise removal and scaling.
9) Got highest predicted value?match the figure with the detected from the neural.

**Experimental Testing Result**

The Table 4.4 depicts the accuracy count for each iteration in the neural network model.Every time training takes place it comes with some upgradation on some models.Here 0-9 has been tested 5 times. Actually 140 datasets have been tested every time. Visualization done by charts in fig 12 Appendix B.

Table 4.4 Experimental Testing Results with the Number of Times Iteration

| Digits | Test 1 (Detected) | Test 2 (Detected) | Test 3 (Detected) | Test 4 (Detected) | Test 5 (Detected) |
|--------|-------------------|-------------------|-------------------|-------------------|-------------------|
| 0 | 60 | 78 | 110 | 116 | 132 |
| 1 | 50 | 43 | 87 | 97 | 124 |
| 2 | 72 | 88 | 86 | 114 | 129 |
| 3 | 23 | 67 | 89 | 109 | 110 |
| 4 | 45 | 46 | 66 | 78 | 125 |
| 5 | 56 | 78 | 68 | 90 | 134 |
| 6 | 76 | 87 | 57 | 104 | 128 |
| 7 | 45 | 40 | 65 | 99 | 138 |
| 8 | 45 | 89 | 95 | 134 | 134 |
| 9 | 55 | 78 | 127 | 130 | 128 |

# CHAPTER 5
# FLOW CHARTS AND FIGURES

**6.1 FLOW CHARTS**

A flowchart is a type of diagram that represents a workflow or process. A flowchart can also be defined as diagrammatic representation of an algorithm, a step by step approach.

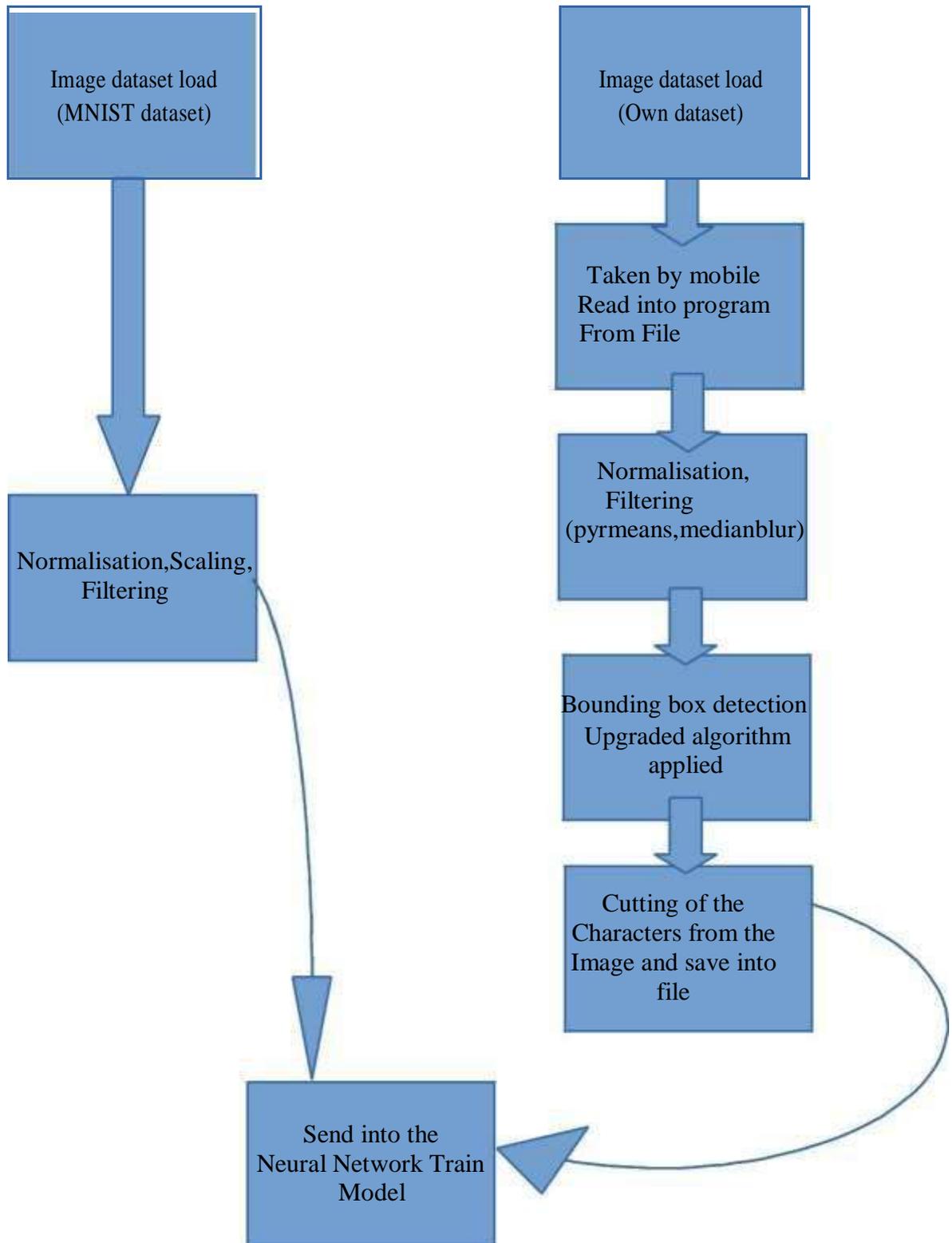

Fig 5.1 Image processing for Training Data

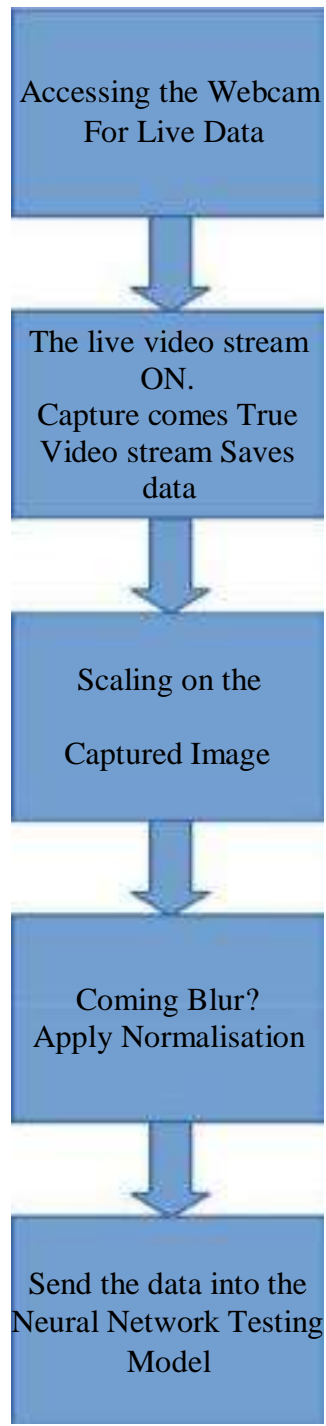

Fig 5.2 Image Processing for Testing data

This Flowchart figure 5.2 shows the Image Processing part For the testing data. Testing data includes the part of the Accessing the webcam and start the video stream on which an image dataset will be tested.

The neural network model includes processing.Various algorithms are depicted here (fig 5.3)

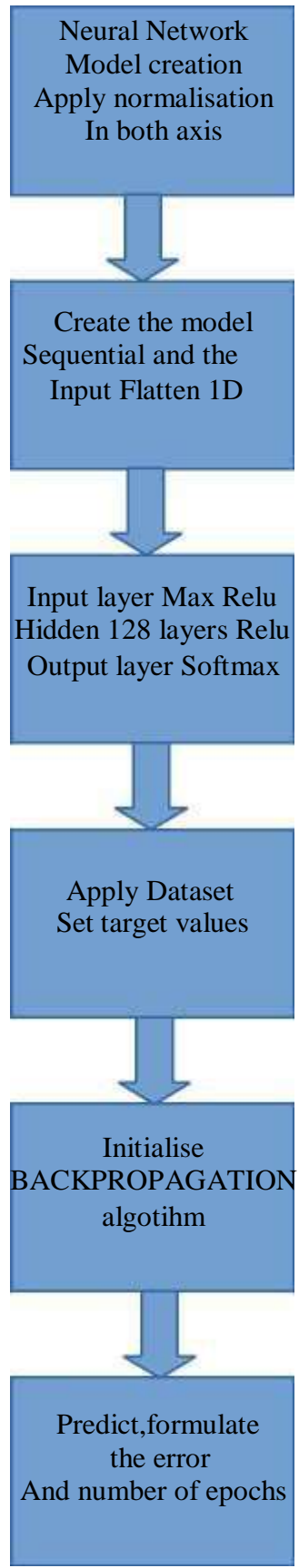

Fig 5.3 Neural Network Model for Training data

The Backpropagation Algorithm (Fig 5.4) applies Stochastic Gradient based algorithm.

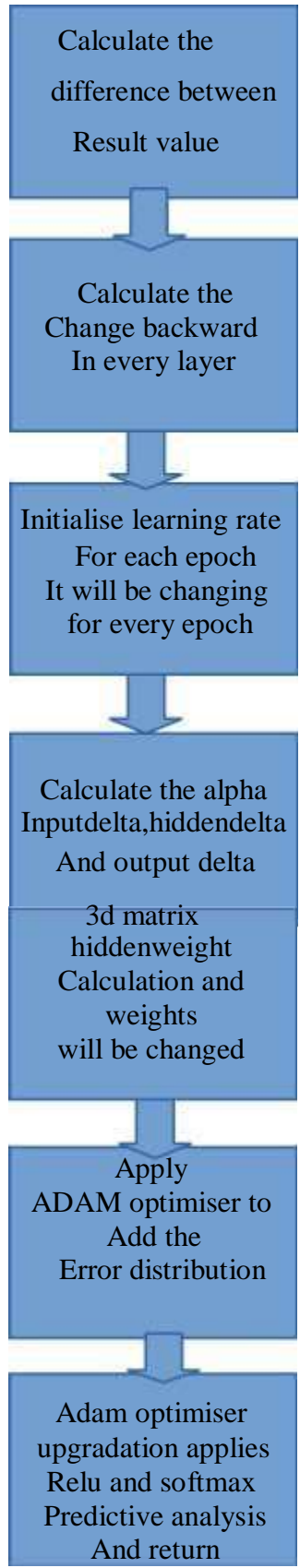

Fig 5.4 Backpropagation Algorithm (upgraded with ADAM optimiser)

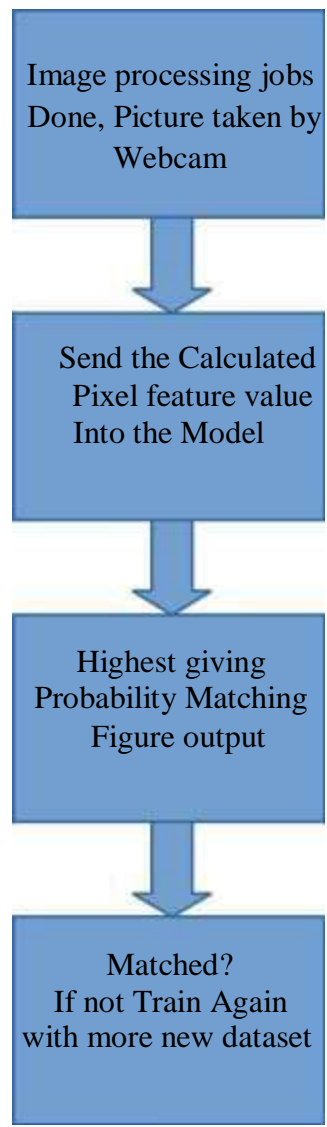

Fig 5.5 Neural Network Model for Testing Data

Neural Network Model gets the prediction from the testing image.Thanks to the Image processing part for the testing image that actually plots the data in the image into the pixels calculated the probabilistic values by sending it into the neural network.If matched then it is a successful one if not then it is an error. May be training will be done with it later. The figure 5.5 depicts the flow of the processes.

# CHAPTER 6

# SYSTEM TESTING

**6.1 TESTING**

Testing is defined as an activity to check whether the actual results match the expected results and to ensure that the software system is defect free. It involves the execution of a software component or system component to evaluate one or more properties of interest. Software testing also helps to identify errors, gaps, or missing requirements in contrary to the actual requirements.

**6.1.1 Unit Testing**

When the testing happens for some individual group or some related units then that type of testing is called as Unit Testing.It is often done by programmer to test the part of the program he or she has implemented.

Unit Testing is successful means all the modules has been successfully tested and it can proceed further.

**6.1.2 Functional Testing**

This type of testing is tested because to check the functional components or the functionality required from the system is gained or not .It actually falls under the testing of the Black Box testing of Software Engineering.This part includes the feeding of the inputs in the system or the project and to check if that system or the project is getting the same value or not as expected if not then calculate the error as wanted and check for more.Functional Testing of this project mainly involves below things. All of these are tested successfully and errors are also calculated.

i)Verifyng the input image
ii)Verifying the work flow
iii)Correct recognition and calculate the error

### 6.1.3 Integration Testing

In a total project or the system, many groups of components are getting added or summed up in the purpose of the project query. Integration testing is about to check the interaction between various modules of the project or the system. This module also includes the hardware and the software requirements of the project.

All the individual modules are integrated and tested together.All the best and extreme cases that the modules are interacting or not are successfully checked and passed,errors are calculated for the machine learning platforms.

### 6.1.4 System Testing

This type of testing is actually meant for the system or the project and also the platform and the integrated softwares and tools,technologies are also tested.The idea or purpose behind the system testing is to check all the requirements that will be provided by the system.

This application of the project along with the tools and technologies has been tested in both windows and linux platform and also unicertified online apple mac platform to check the requirements.It passed successfully.

### 6.1.5 Acceptance Testing

This is a type of system or software testing where a system has been tested for availability.The purpose of this test is to check the business requirements and assess whether it will be accepted for delivery.In this part ADRIAN of pyrimagsearch has been referred to, who worked with the same platform and to check this project accepted by the delivery partner or not.

# CHAPTER 7

# CONCLUSION AND FORESEEABLE ENHANCEMENT

## 7.1 CONCLUSION

This is a project on OCR (Optical Character Recognition).This project is non-fundable project and designed with full of interest which also includes some outer concept on statistical modeling and optimiser technique. In these days of real time analysis, the data has been increased too much(Table 7.1).A small analysis says the size of increase of all of the data in real world.

Table 7.1 Range of Data

| Year Range | Size of data(exabyte) |
|---|---|
| Upto 2005 | 130 |
| 2005-2010 | 1200 |
| 2010-2015 | 7900 |
| 2015-2020 | 40,900 |

Machine learning is an approach to get the real life data into the action over human analysis .This project has an aim to achieve that much goal because all machine learning algorithms intends to go to the better way than a human.

This project is a very much preliminary project based on those .This world entitles the work of Google everyday who himself hasn't achieved that much data also.
This project entitles some different new ideas on

1. Image Processing
2. Machine learning
3. Activation Functions
4. Statistical predictive modeling
5. Optimiser into the programing
6. Text analysis
7. Digit extraction Features.

**7.2 FORESEEABLE ENHANCEMENT**

       This project can be enhanced with a great field of machine learning and arfiticial intelligence. The world can think of a software which can recognise the text from a picture and can show it to the others, for example a the shop name detector. Or this project can be extended to a greater concept of all the character sets in the world. This project has not gone for the total english alphabet becuase there will be more and many more training sets and testing values that the neural network model will not be enough to detect. Think of a AI modeled car sensor going with a direction modeling in the roadside, user shall give only the destination.All of these enhancement is an application of the texture analysis where advanced image processing,Neural network model for training and advanced AI concepts will come.These applications can be modeled further .As this project is fully done by free and available resources and packages this can be also a limitation of the project.

       The fund is very important because all machine learning libraries and advanced packages are not available for free.Unless of those the most of the visualizing platforms like on which developers are doing some works like Watson Studio or Aws.These all are mainly paid platforms where a lot of ML projects are going on.

# CHAPTER 8

# APPENDIX A
# SCREENSHOTS
# INPUTS

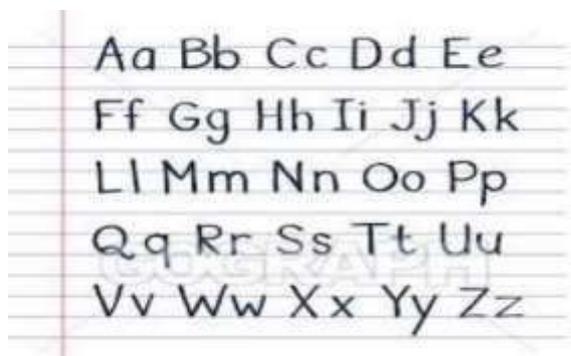

Fig -1 Handwritten Margin pic Input

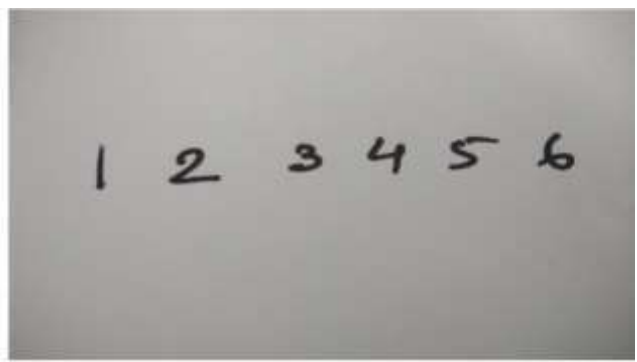

Fig -2 Handwritten Plain paper Input

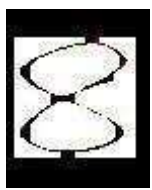

Fig-3 MNIST dataset Input

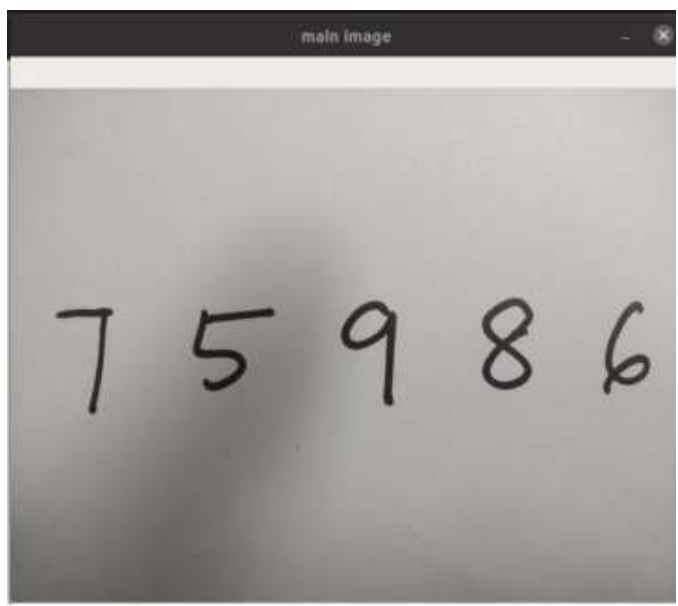

Fig -4 Standard input

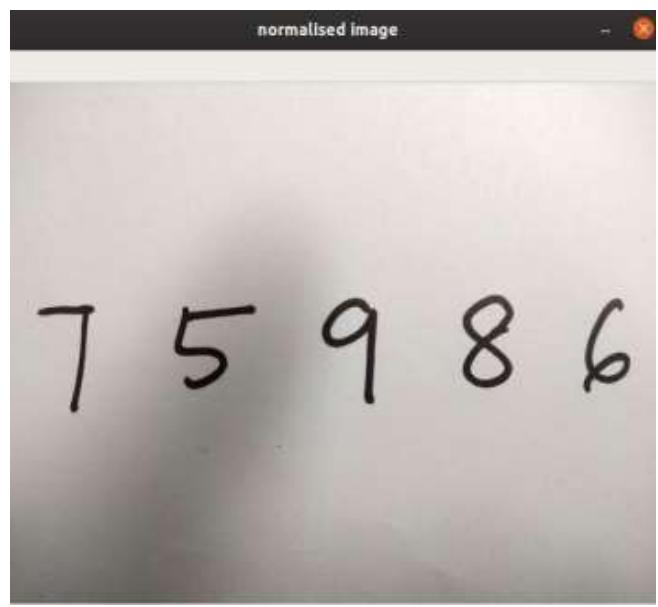

Fig -5 Handwritten Input for
Training after minmax normalisation

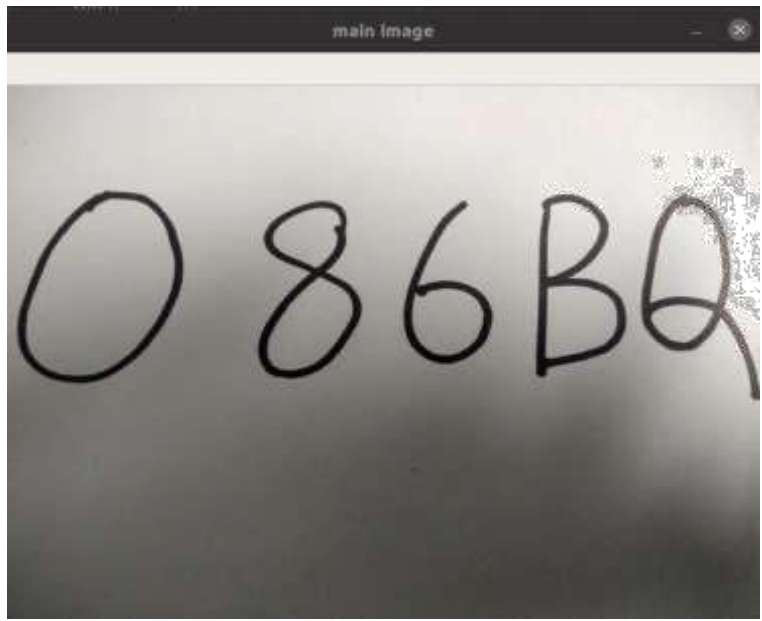

Fig-6 Handwritten Input

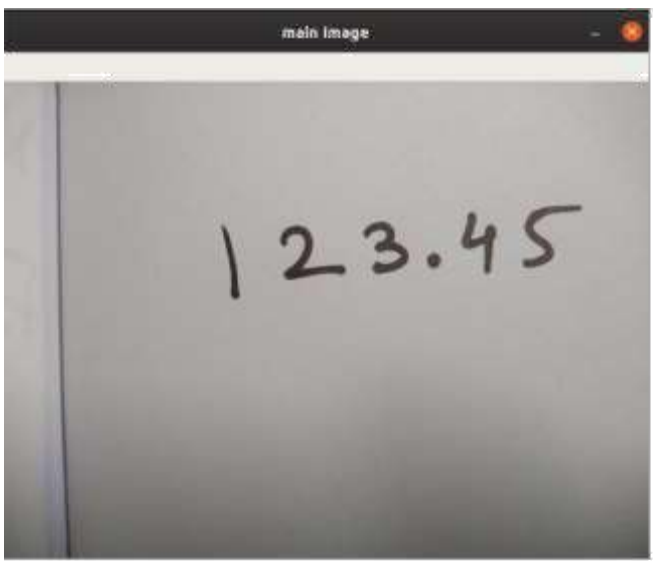 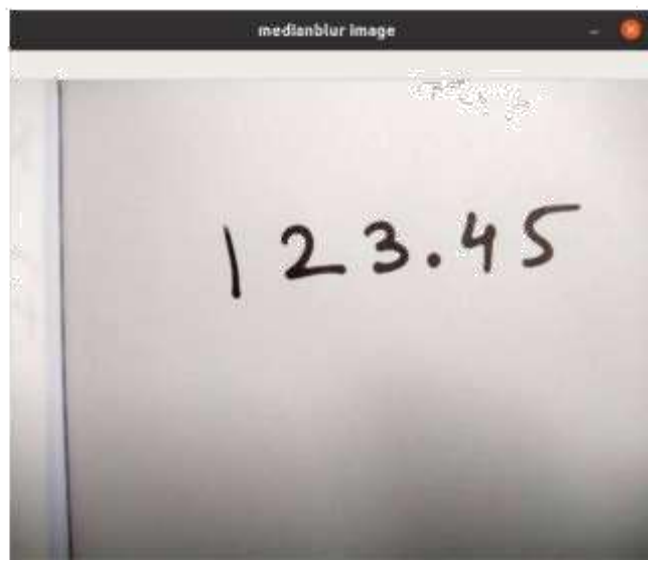

Fig 7 main image                    Fig 8 Median Blur Filter

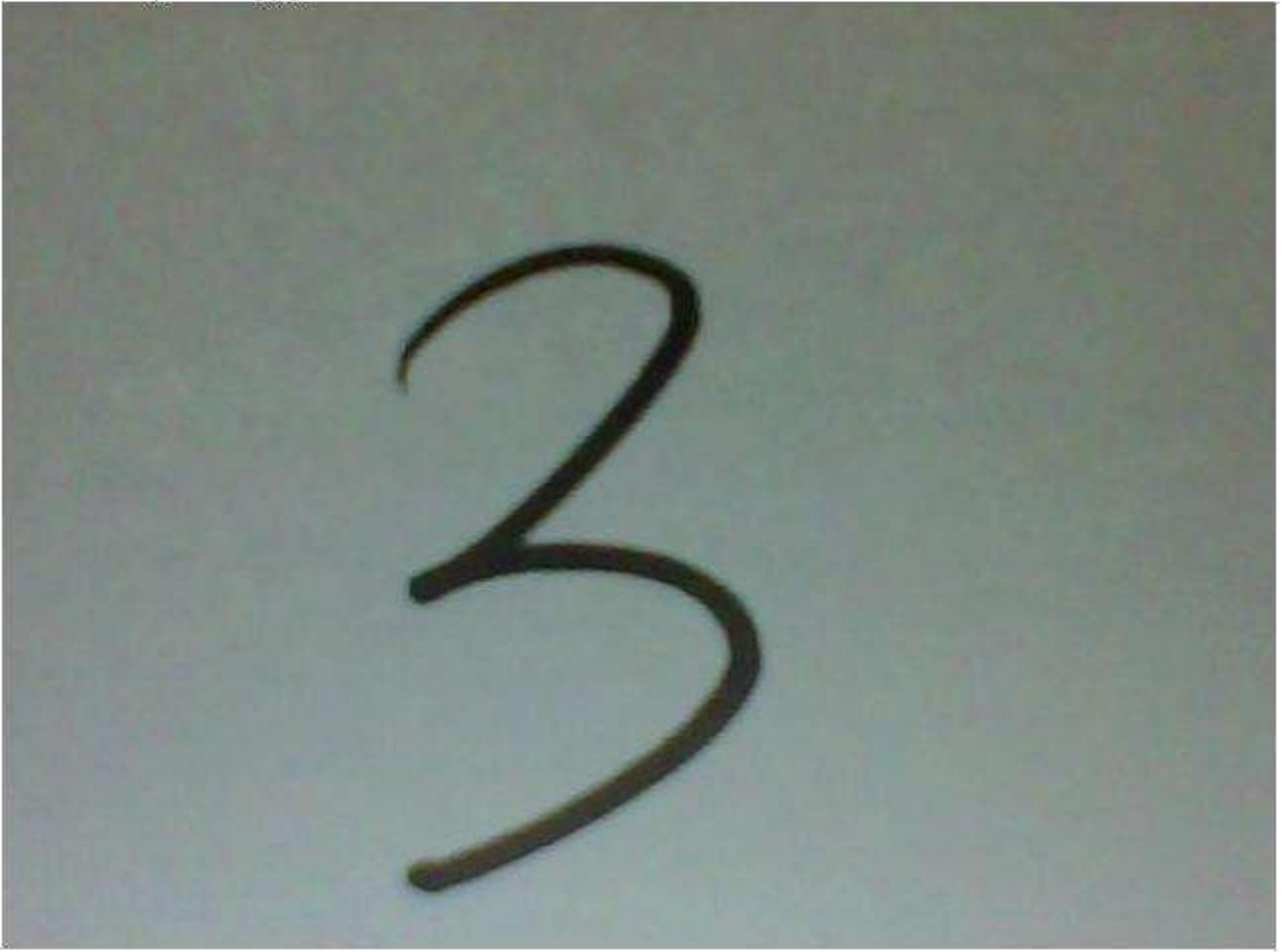

Fig 9 Webcam captured
Image

# APPENDIX B

# OUTPUTS

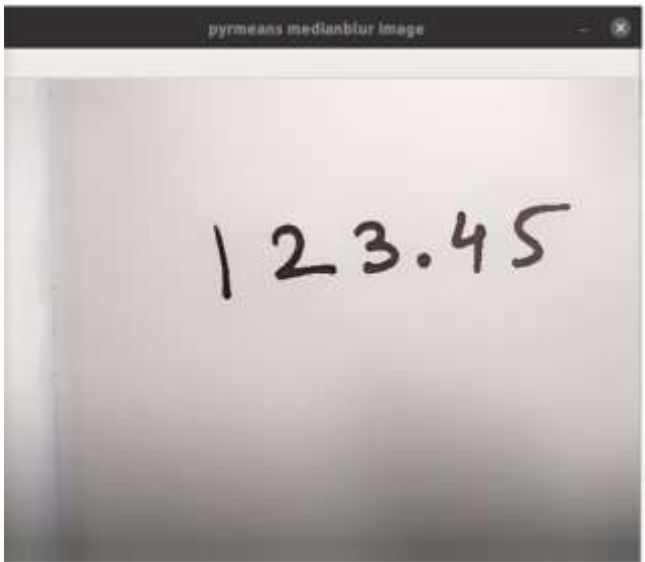

Fig-1 Pyrmeansshift filtering algorithm

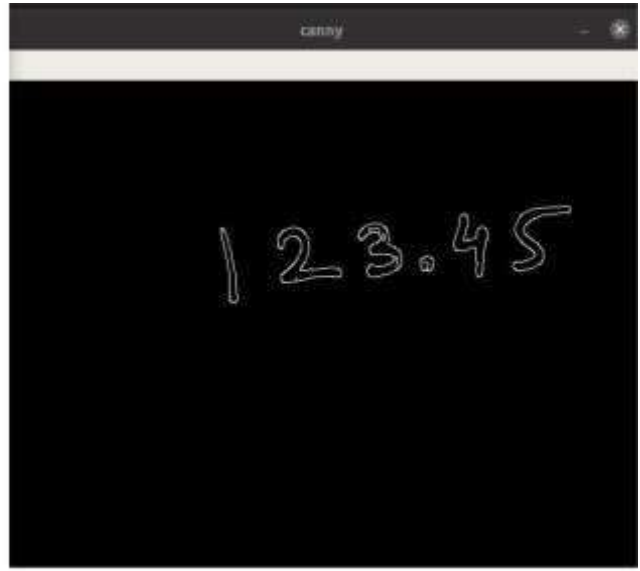

Fig-2 Edge detection

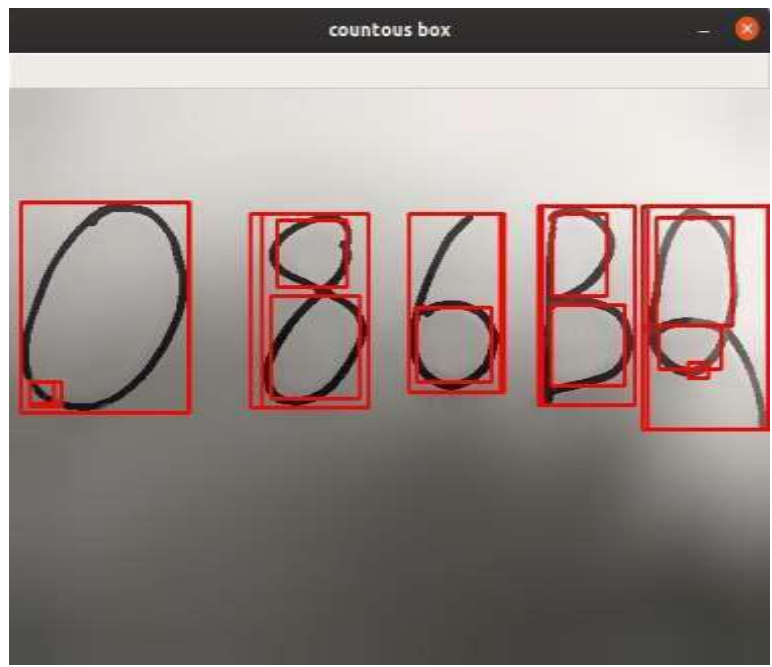

Fig-3 Handwritten input after Normal Contour detection

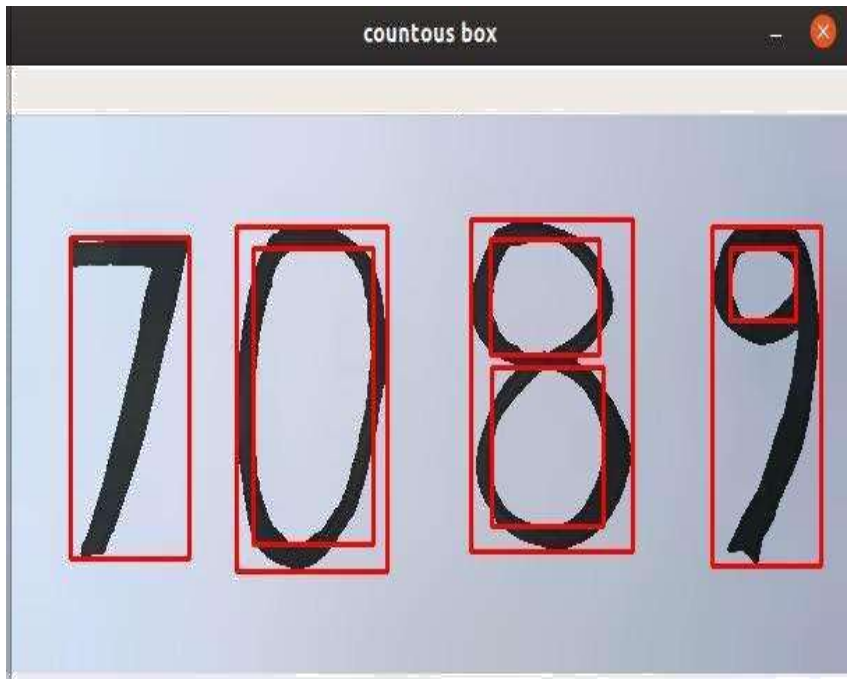

Fig-4 Contour detection algorithm problem

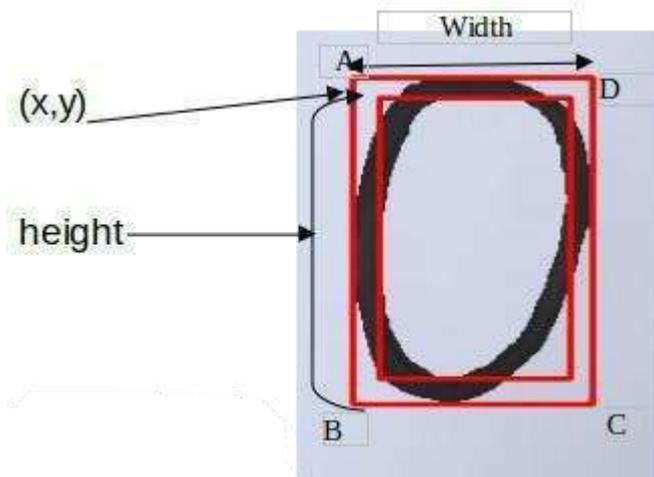

Fig-5 Bounding Box plotting

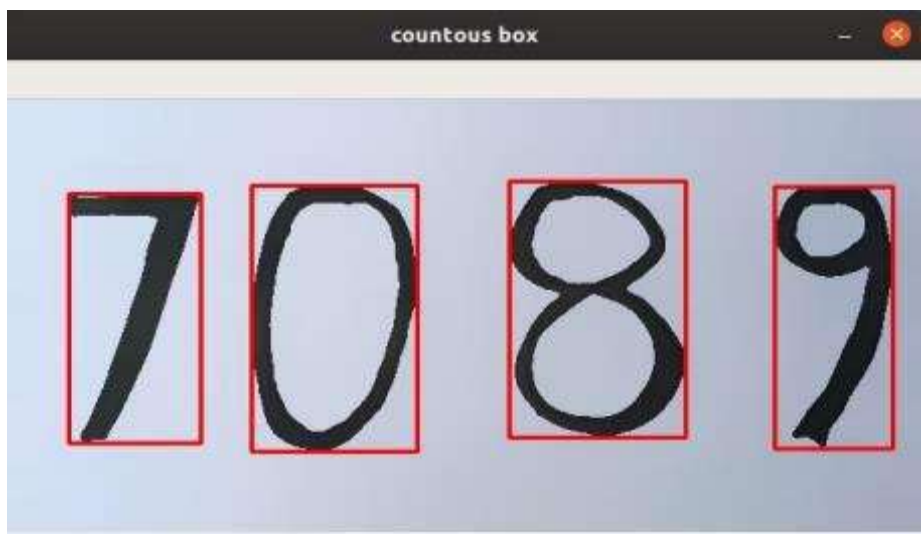

Fig-6 Contour concept upgradation

```
print(np.argmax(predictions[0]))
7

plt.imshow(x_test[0], cmap = plt.cm.binary)
plt.show()
```

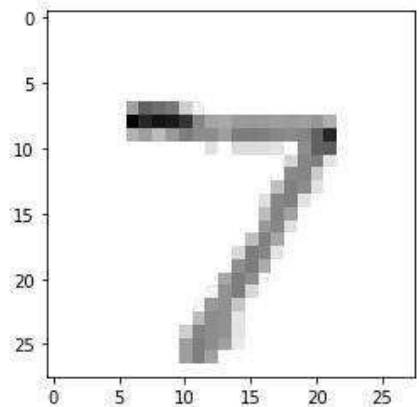

Fig-7 MNIST dataset

```
5s 90us/sample - loss: 0.2646 - acc: 0.9217
4s 62us/sample - loss: 0.1059 - acc: 0.9680
4s 60us/sample - loss: 0.0714 - acc: 0.9769
4s 62us/sample - loss: 0.0525 - acc: 0.9834
4s 62us/sample - loss: 0.0403 - acc: 0.9869
4s 61us/sample - loss: 0.0315 - acc: 0.9897
4s 61us/sample - loss: 0.0266 - acc: 0.9910
4s 61us/sample - loss: 0.0205 - acc: 0.9932
4s 61us/sample - loss: 0.0181 - acc: 0.9938
4s 60us/sample - loss: 0.0160 - acc: 0.9944
```

Fig-8 Training the model

```
print("Probability distribution for a multi-class classification")
predictions[0]

Probability distribution for a multi-class classification
array([6.4432307e-11, 2.9071179e-11, 5.7861965e-10, 4.1306921e-04,
       8.9082035e-14, 8.6323332e-10, 8.8662356e-21, 9.9958664e-01,
       2.0594176e-07, 2.6964958e-08], dtype=float32)
```

Fig-9 Probabilistic value for each set of digits

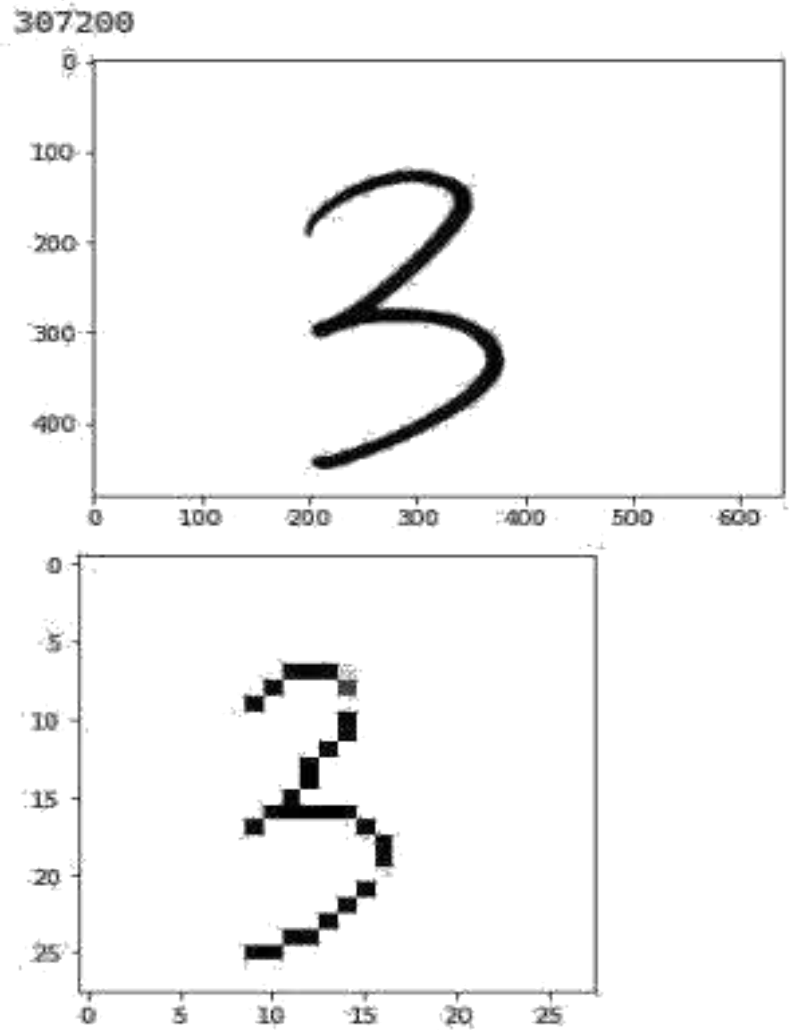

Fig-10 Webcam taken image filtering and scaling

```
Probability Distribution for 0 2.112567e-13
Probability Distribution for 1 3.5835336e-07
Probability Distribution for 2 1.5089857e-10
Probability Distribution for 3 0.9999895
Probability Distribution for 4 3.902672e-14
Probability Distribution for 5 8.723122e-06
Probability Distribution for 6 2.0726507e-17
Probability Distribution for 7 4.961833e-11
Probability Distribution for 8 1.3694355e-06
Probability Distribution for 9 2.605177e-09
The Predicted Value is 3
```

Fig-11 Detection of the image

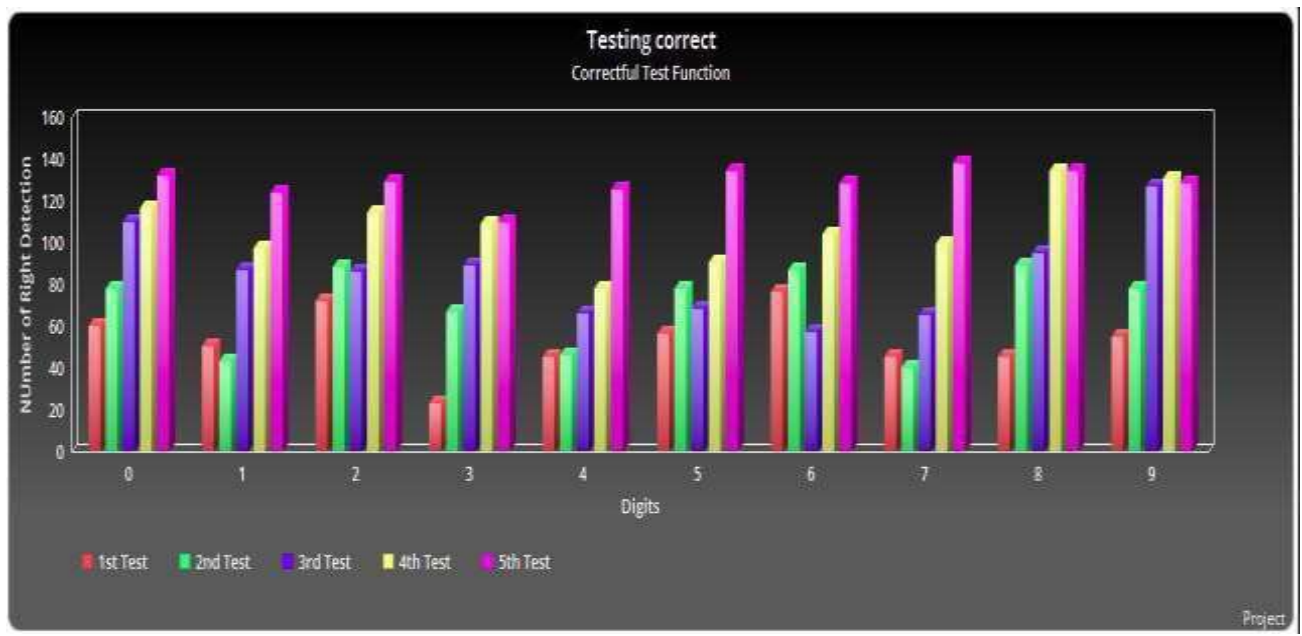

Fig-12 Test dataset result after 5<sup>th</sup> time